\documentclass[letterpaper, 10 pt, conference]{ieeeconf}  

\IEEEoverridecommandlockouts                              

\usepackage{graphicx}
\usepackage{epsfig} 
\usepackage{mathptmx}
\usepackage{amsmath} 
\usepackage{bm}

\usepackage{amssymb}  
\usepackage{amsthm}
\usepackage{wasysym}
\usepackage[mathcal]{euscript}
\usepackage{bbm}
\usepackage{color}
\usepackage{lipsum}
\usepackage[ruled,vlined,linesnumbered]{algorithm2e}
\usepackage[colorlinks=true,linkcolor=black,anchorcolor=black,citecolor=black,filecolor=black,menucolor=black,runcolor=black,urlcolor=black]{hyperref}
\usepackage{etoolbox}
\usepackage[table,xcdraw,dvipsnames]{xcolor}
\usepackage{multirow}
\usepackage{bbm}

\usepackage[flushleft]{threeparttable}  

\usepackage{outlines}
\let\labelindent\relax
\usepackage{enumitem}
\setenumerate[2]{label=\alph*.}
\setenumerate[3]{label=\roman*.}

\makeatletter
\patchcmd{\@makecaption}
  {\scshape}
  {}
  {}
  {}
\makeatother

\newcommand{\regtext}[1]{\mathrm{\textnormal{#1}}}

\newcommand{\tss}[1]{\textsuperscript{#1}}

\usepackage[
    style=ieee,
    doi=false,
    isbn=false,
    url=false,
    eprint=false,
    backend=biber,
    natbib=true
    ]{biblatex}
    
\bibliography{references}

\title{\LARGE \bf
Towards Closing the Loop in Robotic Pollination \\
for Indoor Farming via Autonomous Microscopic Inspection
}

\author{Chuizheng Kong\tss{1}, Alex Qiu*\tss{2}, Idris Wibowo*\tss{1}, Marvin Ren\tss{1}, Aishik Dhori\tss{1}, \\Kai-Shu Ling\tss{3}, Ai-Ping Hu\tss{4}, and Shreyas Kousik\tss{1}
\thanks{\textbf{*} indicates the equal contribution.}
\thanks{$^{1}$ Georgia Institute of Technology, Atlanta, GA.}
\thanks{$^{2}$ Stanford University, Stanford, CA.}
\thanks{$^{3}$ United Stated Department of Agriculture -- Agricultural Research Service, Charleston, SC.}
\thanks{$^{4}$ Georgia Tech Research Institute, Atlanta, GA.}
\thanks{Corresponding author: \texttt{ckong35@gatech.edu}.}
}

\begin{document}

\maketitle
\global\csname @topnum\endcsname 0
\global\csname @botnum\endcsname 0
\thispagestyle{plain}
\pagestyle{plain} 

\begin{abstract}
Effective pollination is a key challenge for indoor farming, since bees struggle to navigate without the sun.
While a variety of robotic system solutions have been proposed, it remains difficult to autonomously check that a flower has been sufficiently pollinated to produce high-quality fruit, which is especially critical for self-pollinating crops such as strawberries.
To this end, this work proposes a novel robotic system for indoor farming.
The proposed hardware combines a 7-degree-of-freedom (DOF) manipulator arm with a custom end-effector, comprised of an endoscope camera, a 2-DOF microscope subsystem, and a custom vibrating pollination tool; this is paired with algorithms to detect and estimate the pose of strawberry flowers, navigate to each flower, pollinate using the tool, and inspect with the microscope.
The key novelty is vibrating the flower from below while simultaneously inspecting with a microscope from above.
Each subsystem is validated via extensive experiments. 
\end{abstract}
\section{Introduction}\label{sec:intro}

\begin{figure}[t]
\centering
\includegraphics[width=1\columnwidth]{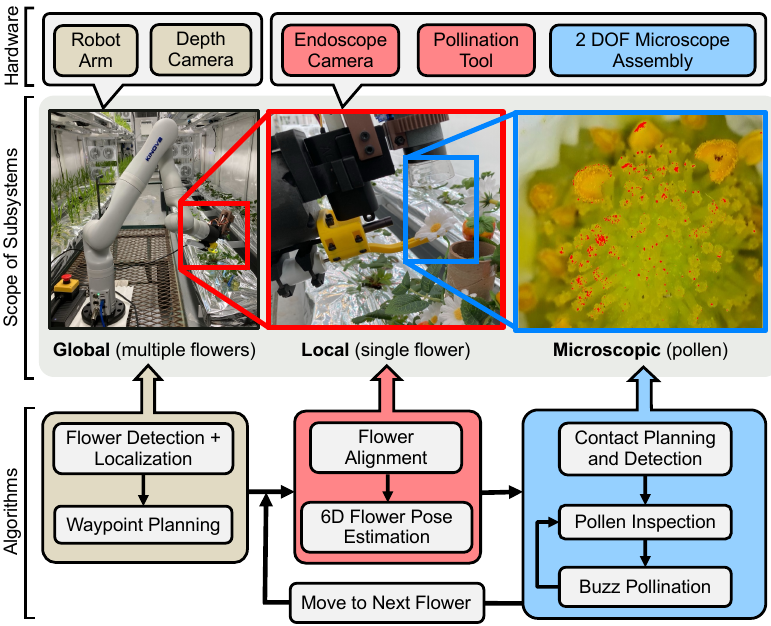}
\vspace{-1em}
\caption{
Our proposed robotic manipulator system performs autonomous pollination and microscopic inspection of strawberry plants as follows.
First, the global scope subsystem estimates flower locations with RGB-D camera.
Then, the local scope subsystem uses a particular flower location as a reference to estimate the 6D flower pose and align the arm for contact.
Next, the microscope subsystem uses visual servoing to facilitate contact with the flower, then alternates between inspection and buzz pollination with our custom, open-source pollination tool. 
Finally, once pollination is confirmed, the robot moves to the next flower.
}
\label{fig: front figure}
\vspace*{-0.5cm}
\end{figure}

Controlled environment agriculture, particularly indoor hydroponic farming in converted shipping containers, has the potential to enable year-round access to locally-grown produce in dense urban areas, especially via small hydroponic facilities.
However, reliable pollination is a major challenge, because bees become disoriented without the sun \cite{dyer1983honey,geiger1995target} and thus struggle to operate indoors, where lighting is consistent in all directions.
The most common existing solution is laborious manual pollination, leading many researchers to explore methods for autonomous robotic pollination \cite{strader2019flower,ohi2018design,nishimoto_evaluation_2023,arugga2024polly,williams2020autonomous,li2022design,broussard2023artificial,smith2024design,dingley2022precision}.



Autonomous robotic pollination poses many challenges.
First, it is critical to robustly detect flowers and estimate their poses despite occlusions from stems and leaves.
Given reliable perception, one must also design a mechanism by which enough pollen is transferred to the female parts of a flower to produce quality fruit without damaging the flower in the process.
Furthermore, it is challenging to verify contact with delicate flowers that may exert too little reaction force to trigger standard tactile sensors.
Finally, it remains open how to autonomously confirm that a flower is sufficiently pollinated to produce a quality fruit without waiting days to weeks for the fruit to develop.
Some crops, like kiwifruit, enable pollen to be sprayed on, enabling autonomous pollination confirmation by adding dye to the pollen \cite{li2022design,williams2020autonomous}.
However, this strategy does not apply to self-pollinating crops, such as strawberries.



\textbf{Contributions:}
Towards solving these challenges, we provide the following contributions:
\begin{enumerate}
    \item an open-source hardware design and integration of a vibrating pollination tool and microscope with an off-the-shelf robotic manipulator arm,
    \item an open-source perception, planning, and control solution for our custom end effector to perform pollination and microscopic inspection, and
    \item extensive experimental validation of each subsystem on both artificial and real strawberry flowers. 
\end{enumerate}
To the best of our knowledge, this is the first approach to perform robotic \textit{in situ} microscopy.
Though this paper focuses on strawberry pollination, we see great promise in using a robot to bring a microscope to samples, as opposed to the traditional approach of bringing samples to a microscope.

The paper is organized as follows:
Section \ref{sec:related_works} reviews related work,
Section \ref{sec:methods} details the proposed method,
Section \ref{sec:validation} describes our experimental system validation, and
Section \ref{sec:conclusion} provides concluding remarks and future directions.

\section{Related Work}\label{sec:related_works}

We now review concepts related to our system and discuss what scope (global, local, micro) each concept is in. 

\subsubsection*{Artificial Pollination (Global)}\label{subsec:related_work_pollination_mech}
For self-pollinating crops like strawberries, pollen must be moved from male to female parts of a flower.
Common artificial pollination approaches include air, brush, and buzz strategies.
Air pollination uses a contactless burst of air to dislodge pollen, which is effective for crops such as tomatoes \cite{arugga2024polly}.
Brush pollination involves softly contacting the flower head, which has been achieved robotically with end effectors inspired by bee physiology \cite{nishimoto_evaluation_2023,ohi2018design}.
Buzz pollination is similarly inspired by the vibration caused by a bee \cite{vallejo2019buzz}, and has been replicated robotically \cite{tayal_efficiency_2020}.
We provide a unique approach to robotic buzz pollination from \textit{beneath} the flower to allow inspection from above.

\subsubsection*{Evaluating Pollination (All Scopes)}
For self-pollinating crops, underpollinated flowers lead to undesirable fruits (we confirm this experimentally; see Tab.~\ref{table:pollination_result} and Fig.~\ref{fig:pollination_exp}).
However, overpollination can also be harmful \cite{macinnis2017quantifying}.
This leads to the question, how can one best evaluate pollination?
The typical strategy is indirect, based on fruit yield and quality between a pollinated test group and an unpollinated control group \cite{nishimoto_evaluation_2023}.
To the best of our knowledge, most robotic pollination devices employ this indirect evaluation strategy \cite{ohi2018design, strader2019flower}.
Another strategy is to remove flowers and observe them under a microscope \cite{symington2024strawberry}, though this is clearly destructive.
Luckily, pollen deposition can be directly measured \textit{in situ} with macro photography \cite{macinnis2017quantifying}, though this has not been carried out robotically.
We address the above gaps with an autonomous robotic microscopic system for direct evaluation.
Note that, for crops with separate male and female parts (e.g., kiwifruit), robotic spray pollination is possible to directly evaluate \cite{williams2020autonomous,gao2023novel}.

\subsubsection*{Flower Detection and Pose Estimation (Local)}
Reliably finding flowers remains a challenge in agricultural robotics; we find that this is the most common point of failure in our own method.
Most strategies involve deep-learning-based classifiers \cite{strader2019flower,YANG2023,ahmad2024}.
Another strategy for sweet pepper fruit builds a 3-D model, then fits an ellipsoid to the surface.
We attempt to make the best of both worlds by using both a deep-learning classifier \cite{yolo} and a 3-D fit \cite{RAFT2020}.


\subsubsection*{Manipulator Control for Pollination (Local)}
Most state-of-the-art robot arm pollination systems use a single point of contact from above the flower stigma \cite{nishimoto_evaluation_2023, wei2024novel, strader2019flower}.
The associated manipulation problem is often straightforward once the on-board camera solves for precise pose estimation of the flowers. 

However, our proposed pollination method involves more complex contact modes such as target aligning and capturing, which can sometimes be found in harvesting mechanism designs.
\cite{xiong2020autonomous} developed a customized end-effector with built-in infrared sensors and a engulfing gripper design for close-loop strawberry harvesting.
\cite{qiu2023tendon} implemented a berry harvesting gripping the involves eye-in-hand close-loop alignment when visiting each berry to be harvested. 
Our manipulation control takes inspiration from both works above, using two eye-in-hand camera (endoscope and microscope) for precise flower alignment and an engulfing mechanism design to stabilize the flower during the pollination and pollen evaluation process.

\subsubsection*{Mobile Autofocusing Microscopy (Micro)}
There are a variety of strategies for autofocusing benchtop microscopes \cite{pech-pacheco2000diatomautofocus, sun2005afalgoselect, pertuz2013focusmethod, zhang2024autofocusmethod}, and robotic actuation to move samples within the microscope view under human direction \cite{andreyev2024samplehold}.
To the best of our knowledge, all of these systems require bringing a sample to the microscope; we flip this paradigm by bringing the microscope to the sample, a strategy which is common and useful in surgery \cite{antoni2015robomicgesture,montemurro2022exoscope}.
Our use case is inspired by using a macro camera to assess pollination directly \cite{macinnis2017quantifying}.
An alternative to a macro camera is a handheld, mobile microscope, but these are challenging to use due to the need for a careful, steady grip to manually focus.
We address this challenge with an autonomous robot to steer and autofocus the microscope.

\begin{figure*}[h!]
\centering
\includegraphics[width=1\textwidth]{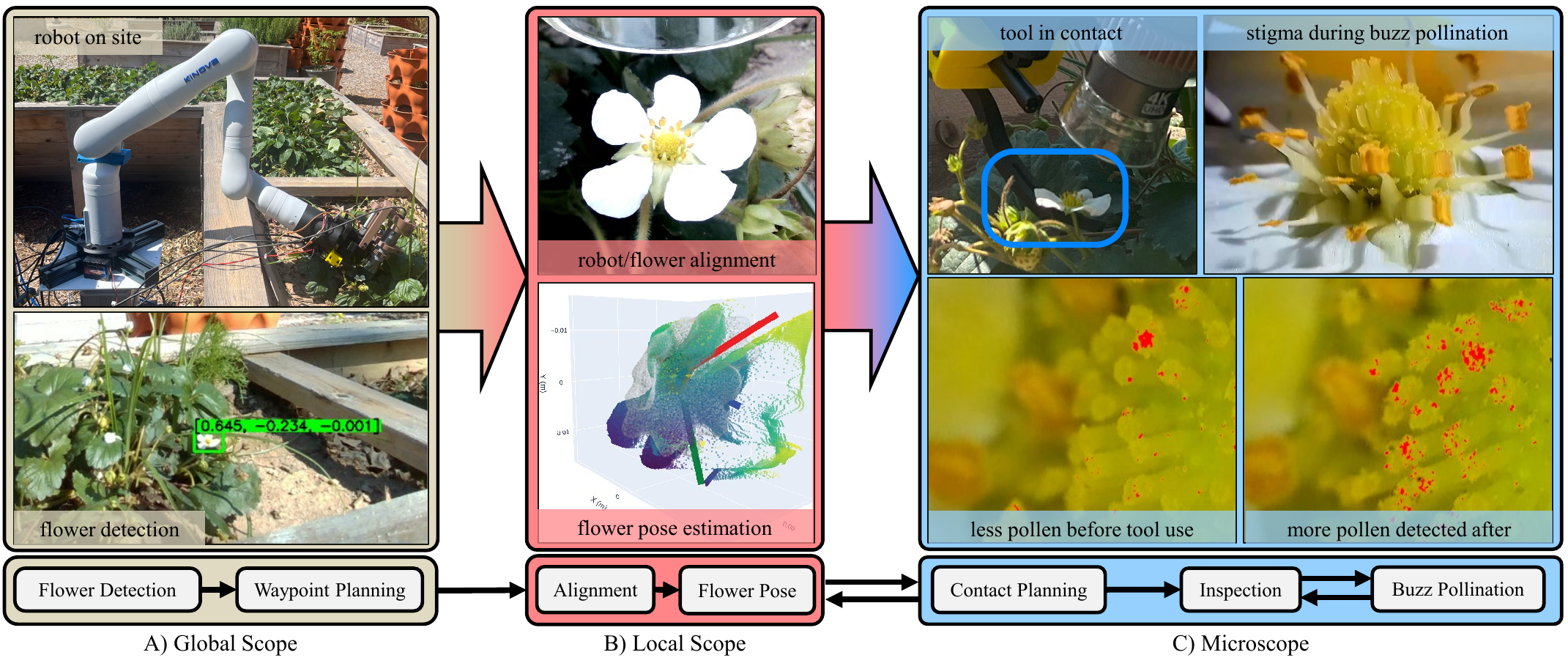}
\vspace*{-0.4cm}
\caption{
Overview of our proposed autonomous robotic pollination system.
}
\label{fig:method_figure}
\vspace*{-0.5cm}
\end{figure*}

\section{Methods} \label{sec:methods}
We propose a system for closing the loop in robotic pollination, from global scope to local scope to microscope (see Figs.~\ref{fig: front figure} and \ref{fig:method_figure}).
Our code and hardware designs are open source \url{https://github.com/kczttm/IndoorFarming}.

\subsection{Global Scope}
This subsystem detects flowers $\approx1000$ mm away, then steers our custom end effector so that it can visually servo to each flower.

\subsubsection{Hardware}
In the global scope, we use a Kinova Gen3 7-DOF robotic arm and an Intel RealSense D435F depth camera rigidly mounted to the robot's first link.
The arm is equipped with a custom end effector containing a buzz pollination tool, endoscope (i.e., local scope camera), and microscope with autofocus assembly.
We use ROS2 \cite{macenski2022ros2} to communicate, and a laptop with an Nvidia RTX 4090 GPU for computation.

\subsubsection{Flower Detection and Localization}\label{subsubsec: global flower detection}
To detect strawberry flowers, we use YOLOv8 \cite{yolo} fine-tuned on the Roboflow dataset \cite{strawberry-flower_dataset}.
We found that strawberry flowers are difficult to sense $> 500$ mm away, because such images are not present in the dataset.
To mitigate this, we apply \cite{akyon2022sahi} to cut the image into smaller pieces before running YOLOv8.
We localize each flower by masking the RealSense depth estimate with the YOLOv8 bounding box to get the flower center in the RealSense local frame, then using the camera's forward kinematics to transform to the robot's baselink (global) frame.

\subsubsection{Waypoint Planning}\label{subsbsec: global waypoint planning}
The flower center location provided above are usually not precise enough for pollination contact. 
Our solution is to enable our end effector, equipped with an endoscope (detailed below) to visually servo towards the flower.
We found that a surprisingly simple heuristic is effective to prepare for such action.
Given a flower location in the baselink frame, we create a line segment between the RealSense camera and the center of the flower.
We create a 3-D position waypoint on that line segment at $60$\% of the way from the RealSense to the flower; we found that the range of $60-80$\% worked best in practice (note that the flowers are usually $1000$ mm from the realsense). 
We then navigate the endoscope onto this waypoint while achieving a desired 3-D orientation by aligning the endoscope with the line segment.
The motivation is that the flower should be within the field of view of the endoscope camera to enable the local scope subsystem to operate.
We use the Kinova Gen3's default position control to track the waypoints and orientations for each of the localized flowers.

\subsection{Local Scope Subsystem}\label{subsec: local scope subsystem}
This subsystem is responsible for locally steering our pollination tool to a given flower.

\subsubsection{Hardware}
\begin{figure}[!b]
\vspace*{-0.6cm}
\centering
\includegraphics[width=1\columnwidth]{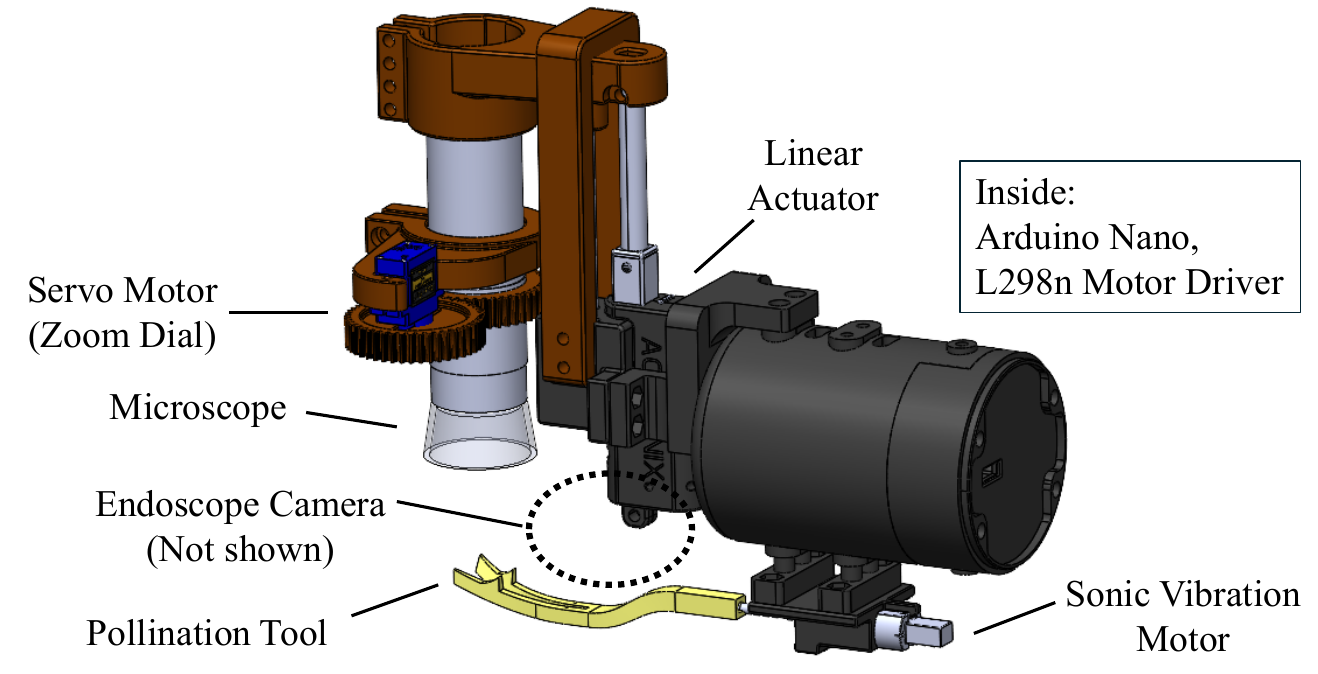}
\caption{An overview of our custom end effector.}
\label{fig: microscope subassembly}
\end{figure}

Though we assume the RealSense camera's pose is known in the baselink frame, we found that even small angular error $(\approx0.05^\circ)$ in its pose can cause up to $50$ mm of location estimation error for flowers that are $1000$ mm away from the camera, which is larger than the diameter of most strawberry flowers of $22-34$ mm \cite{symington2024strawberry}.
We solve this issue by enabling close-range visual servoing using a monocular 2K RGB USB endoscope camera with a $70 - 400$ mm focal range.
We mount our custom pollination tool directly under and aligned with the endoscope.
These components and their assembly are shown in Fig.~\ref{fig: microscope subassembly}.

\subsubsection{Pollination Tool Design}\label{subsubsec: pollination tool design}

\begin{figure}[t]
\centering
\includegraphics[width=1\columnwidth]{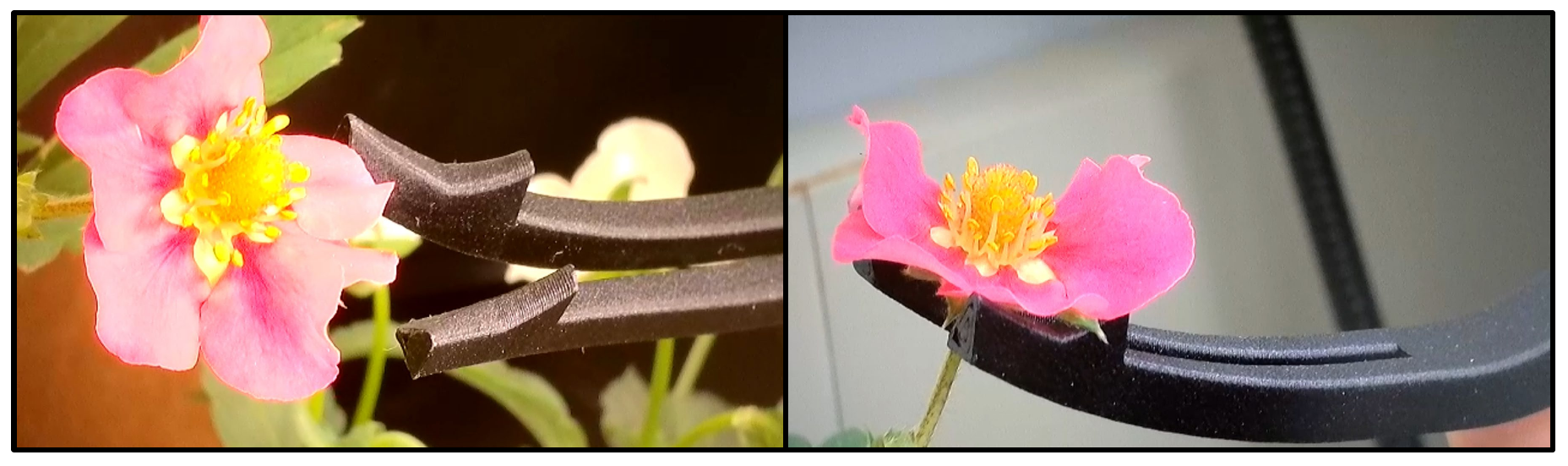}
\vspace*{-0.5cm}
\caption{Close-up view of our pollination tool before and during contact with a real strawberry flower.
The cup shapes on the tines at the distal end of the tool help to grasp the flower.}
\label{fig: pollination fork}
\vspace*{-0.5cm}
\end{figure}

We propose a buzz pollination tool to contact the stem below the flower, enabling simultaneous microscope inspection from above.
The tool's distal end is shown in Fig.~\ref{fig: pollination fork}.
Preliminary testing revealed that several key features were needed.
First, the tool is forked to trap the stem and increase robustness to flower pose error.
Second, the tool is curved to tension the flower stem and better transmit vibration.
Third, an inverted cup structure on the upper face pushes the petals towards each other to prevent dispersion of pollen away from the flower during vibration.
Furthermore, since the flower is between the tool and the microscope, we use the microscope cover the flower and further prevent dispersion.
Finally, the tool is attached to a sonic vibration motor that oscillates at $\approx 250$ Hz.
To prevent damaging the flower and stem, the tool is shaped so no part moves more than $\approx 1$ mm during vibration.

\subsubsection{Flower Alignment via Visual Servoing}\label{subsubsec: local flower alignment}
Using the endoscope image, our goal is to place the endoscope camera and tool directly in front of the flower of interest.
Since our endoscope is already aligned with the flower from the global scope, we modify the classical image-based visual servoing approach \cite{hutchinson1996tutorial} to only control the endoscope translation.
Depending on the endoscope location and density of flowers on a plant, many flowers may be present in the endoscope image and detected by YOLOv8; we servo based on the flower closest to the center of the image.
The endoscope has no depth information, so we use a heuristic based on average strawberry flower diameter \cite{symington2024strawberry} to approximate the diagonal of the bounding box as $24\sqrt{2} \approx 33$ mm.
The Kinova Gen3 accepts end effector translation velocity commands, so we use PD control to perform servoing; we set desired velocity to $0$ and desired position $100$ mm away from the endoscope camera frame origin in the direction of the line segment output between the RealSense and the flower.

\subsubsection{6-D Flower Pose Estimation}\label{subsubsec: local pose estimation}
After flower alignment, we need to align the pollination tool to make contact.
The point cloud resolution of the RealSense camera is not sufficient for pose estimation of individual flowers.
Instead, we take two endoscope images $5$ mm apart using the arm's end-effector translation controller, then compute per-pixel optical flow with the RAFT model \cite{RAFT2020} to get a local depth estimate and point cloud.
We again run YOLOv8 to get a flower bounding box as a mask to extract the flower point cloud.
We then use the Random Sample Consensus (RANSAC) \cite{fischler1981random} and Iterative Closest Point (ICP) algorithms via Open 3-D \cite{open3d} to match the flower to a template flower point cloud.
The result of this process is an outward facing normal vector from the flower center, which we use for pollination tool contact and microscopic inspection.

\subsection{Microscopic Subsystem}\label{subsec: microscope subsystem}
Our smallest-scope subsystem uses a microscope to confirm tool contact and quantify pollen deposition.

\subsubsection{Hardware}\label{subsubsec: microscope hardware}
We use a hand held monocular 2k RGB USB microscope with $50-1000\times$ magnification.
The microscope lens is ringed by white LEDs and is centered in a cup that we use to catch and stabilize flowers.
The microscope is mounted on a linear actuator, and facing in towards the cup of the pollination tool for movement onto a trapped flower; it can move between $0-45$ mm from the tool.
The manual zoom wheel of the microscope is actuated by a servo motor.
Both the linear actuator and servo motor are controlled by an Arduino Nano board with the pySerial library.

\subsubsection{Contact Planning and Detection}\label{subsubsec: contact planning and detection}
We use a heuristic flower contact strategy which we found reduces the total end effector motion, and thereby limits the chance of plant damage or collision with hydroponic shelving in an indoor farm (we saw no damage or collisions in our testing).
First, we rotate to align the pollination tool to be perpendicular to the flower outward normal vector from RANSAC/ICP, since this is the most likely direction of the flower stem.
Second, we lower the tool in the flower stem direction until the upper tool surface is lower than the lowest point of the flower point cloud, to prevent the tool from hitting the flower as it approaches.
Third, we drive the tool forward until the estimated flower center is above the center of the tool's cup.
All of this motion is done with the linear actuator at $45$ mm from the tool, and the zoom dial such that the flower is likely to be in focus when contacted by the tool, based on calibration with an average-sized flower.

The previous step leaves the flower approximately under the microscope, which we use to close the loop for centering the flower.
Note, the microscope image is almost always out of focus at this stage.
We use an OpenCV Hue/Saturation/Value (HSV) filter \cite{itseez2015opencv} to find a contour of the flower center, then fit an ellipse on the contour to estimate the flower center location and the size of the flower stigma.
The filter has heuristically-determined lower and upper HSV bounds of $[20, 215, 100]$ and $[30, 255, 245]$.
Once the flower center is located, we translate the tool ``up'' in the stem direction while also adjusting it via translation perpendicular to the stem direction to place the flower center in the middle of the microscope image with proportional controller.
We stop raising the tool when the flower comes into focus, using a focus score (discussed next) on pixels in the bounding box of the flower center ellipse.
Finally, we lower the microscope's linear actuator to a predetermined, fixed distance of $6$ mm away from the fork tip, which we found typically captures a flower in the microscope's cup.

\begin{figure} \label{fig:focus_score}
    \centering
    \includegraphics[width=0.9\linewidth]{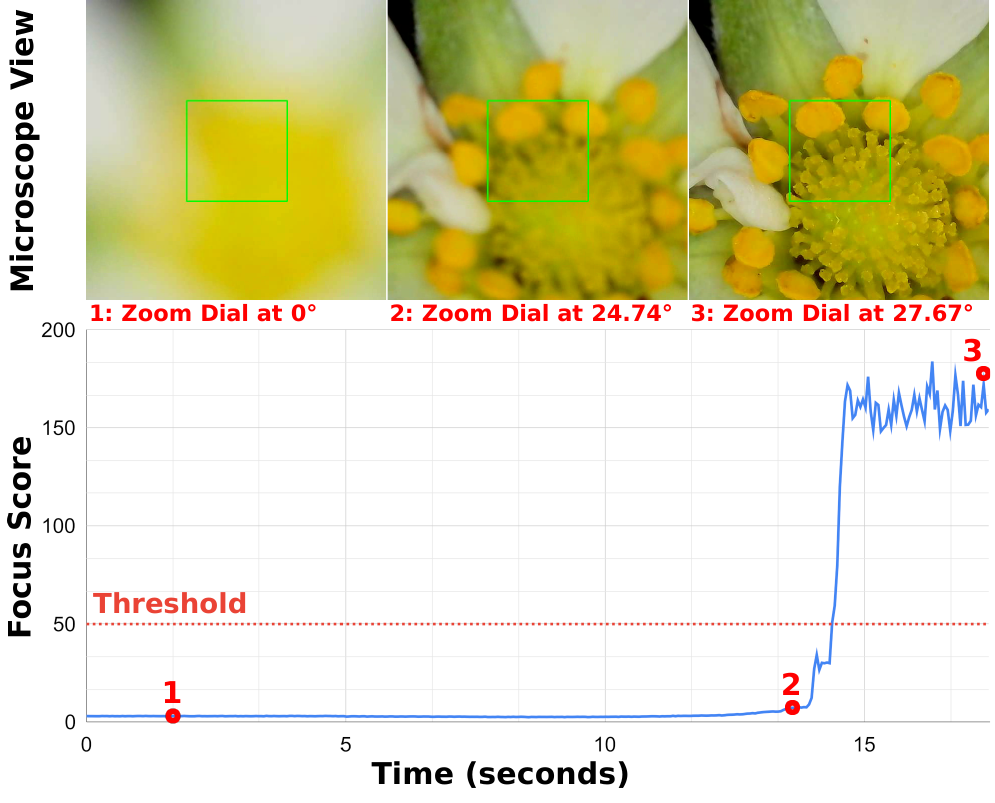}
    \vspace*{-0.3cm}
    \caption{Visualization of autofocus operation, which stops once the focus score surpasses the threshold.
    Time points 1, 2, and 3 match to the microscope views at the top.}
    \vspace*{-0.3cm}
\end{figure}

\subsubsection{Microscope Focus Score and Autofocus}\label{subsubsec: microscope focus score and autofocus}
We implement a focus score that adapts existing work \cite{pech-pacheco2000diatomautofocus, sun2005afalgoselect, pertuz2013focusmethod}.
In particular, we compute the variance of the image Laplacian using a kernel
\begin{equation} \label{eq:laplkern}
    K = \left[\begin{smallmatrix}
        0 & 1 & 0\\
        1 & -4 & 1\\
        0 & 1 & 0
    \end{smallmatrix}\right].
\end{equation}
We implement this in OpenCV \cite{opencv_library} with \texttt{ksize=1}.
Note, we only apply this to pixels in the bounding box of the flower center ellipse from above.
Unlike other autofocus approaches \cite{pertuz2013focusmethod}, we do not apply Gaussian blur or grayscale conversion first, because we found these reduced the focus score performance.
The focus score typically varies from $0$ to $1000$, with values above $50$ in focus (see Fig.~\ref{fig:focus_score}).
To bring the microscope image into focus, we use proportional control to actuate the zoom dial (via the servo motor).

\subsubsection{Pollen Deposition Inspection}\label{subsubsec: pollen deposition inspection}
Once contact is confirmed and autofocus is complete we inspect pollen deposition using two HSV filters in ImageJ \cite{abramoff2004image}, following \cite{macinnis2017quantifying}.
First, we remove the background using lower and upper HSV bounds of $[0,170,0]$ and $[45,255,255]$.
Then we isolate the pollen with lower and upper HSV bounds of $[0,0,154]$ and $[62,168,255]$.
This highlights pollen in red as in Fig.~\ref{fig:method_figure}(C).
The number of red pixels is our approximate pollen count.

\subsubsection{Pollination/Inspection Loop}\label{subsubsec: pollination inspection loop}
After the initial inspection,  we alternate between vibrating the flower for a fixed duration ($10$ s in our experiments) and redoing the inspection until the pollen count exceeds a threshold.
We found that, before pollination, the pollen count is on the order of $10^3$; after pollination it is on the order of $10^5$.

\section{Experimental System Validation}\label{sec:validation}

\begin{figure}
    \centering
    \includegraphics[width=1\linewidth]{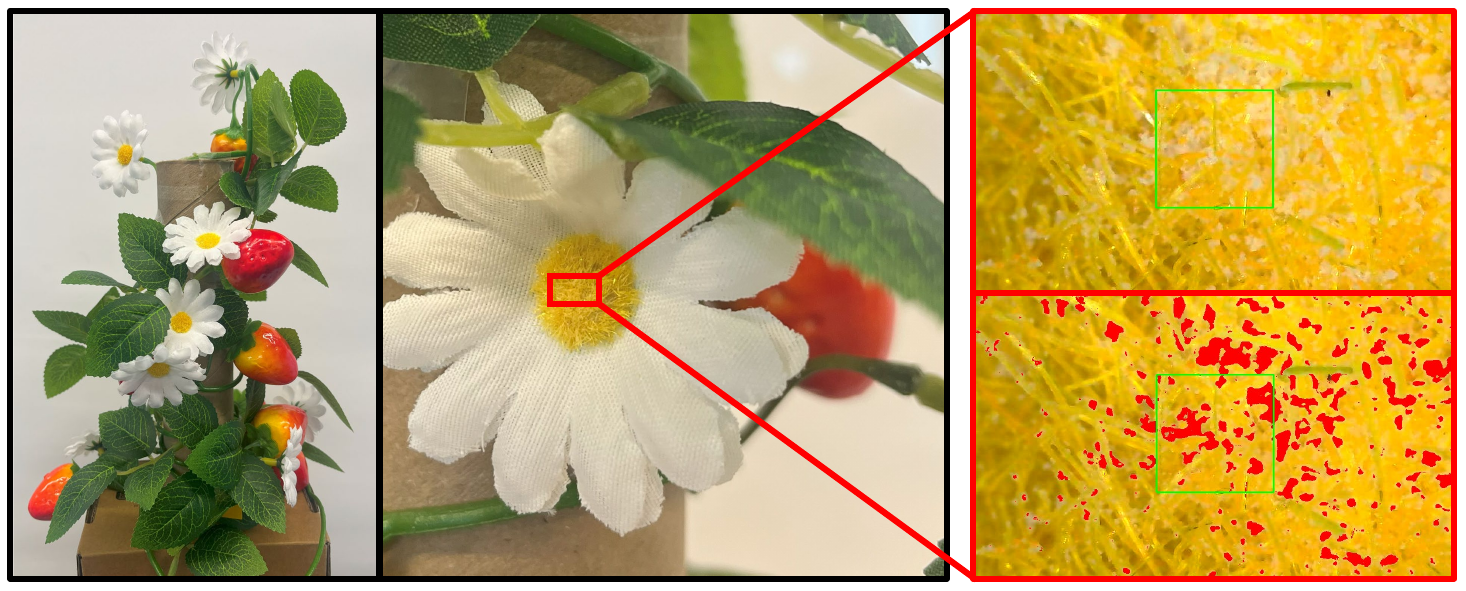}
    \vspace*{-0.3cm}
    \caption{(Left) The artificial strawberry flower plant used for system validation.
    (Center) An example of a flower with partial occlusion.
    (Right) The use of talc powder as artificial pollen.
    Note that artificial pollen is present in the upper right figure; it is autonomously detected and marked red with the approach in Sec.~\ref{subsubsec: pollen deposition inspection}.}
    \label{fig: artificial plant and flower pollination detail}
    \vspace*{-0.5cm}
\end{figure}

We now experimentally validate each subsystem by deploying the entire system:
\begin{itemize}
    \item We test the global flower detection, local flower approach and pose estimation, and pollination tool contact via full system deployment (i.e., each subsystem proceeded based on the previous subsystem's success).
    At the time of testing (9/2024), the strabwerry flowering window in Atlanta, GA had already passed, so we used the artificial strawberry plant shown in Fig. \ref{fig: artificial plant and flower pollination detail}.
    The plant has 8 adjustable flowers, allowing us to replicate difficult scenarios such as flower clustering and occlusions.
    We summarize our results in Tab. \ref{table: autonomous pipeline full results}.
    
    \item We separately tested our microscopic inspection system on artificial flowers due to the need to apply artificial pollen by hand.
    Results are in Sec.~\ref{subsubsec: pollination inspection experiment}.

    \item We also tested the pollination tool separately in a USDA indoor controlled environment facility on real flowers (pollinated by hand).
    Results are in Tab.~\ref{table:pollination_result} and Fig.~\ref{fig:pollination_exp}.
\end{itemize}

\subsection{Global Scope Subsystem Validation}

The global scope detects flowers (discussed here) and plans waypoints for each flower (discussed in local scope).

\subsubsection{Flower Detection and Localization}
We hypothesis that our RealSense camera paired with YOLOv8 will detect and localize all flowers facing the camera (i.e., flower center visible) within $1,000$ mm and in its field of vision. 

\textit{Experiment Setup:}
We placed an artificial strawberry plant at 20 different positions and rotations on a table approximately $1000$ mm in front of our robot indoors under fluorescent lighting, resulting in a total of 50 visible flowers.

\textit{Results and Discussion:}
The system successfully detected 40/50 (80\%) of the flowers.
Of these, all 40 were localized correctly.
Our YOLO model was trained on an online, open-source data set, as opposed to images taken from our robot, so we believe this result can be significantly improved with fine tuning on a custom dataset.

\subsection{Local Scope Subsystem Validation}
\subsubsection{Flower Alignment}
We hypothesize that this subsystem can align with all flowers detected by the global subsystem, meaning centering the endoscope in front of each flower.

\textit{Experiment Setup:}
For each flower, the robot tracks the waypoints from Sec.~\ref{subsbsec: global waypoint planning}, then outputs true/false if pose error of the flower's YOLO bounding box falls below a set threshold (i.e., autonomously confirms success).

\textit{Results and Discussion:}
The subsystem successfully aligned with 38/40 (95\%) of the flowers.
For the two failed cases, the flower was visible to the RealSense camera but became occluded by a leaf during robot motion.
Future work will consider obstacle avoidance and plant deformation to mitigate this issue.

\subsubsection{6-D Flower Pose Estimation}
Given a flower centered in front of the endoscope, we hypothesize that this subsystem will provide a pose correctly aligned with the flower/stem.

\textit{Experiment Setup:}
Starting from the front of each flower, the robot takes two endoscope images $5$ mm apart, constructs a point cloud, then performs point cloud registration with RANSAC/ICP as in Sec.~\ref{subsubsec: local pose estimation}.
Success is measured autonomous as the point cloud registration mean squared error falling below a threshold value.

\textit{Results and Discussion:}
Excitingly, 38/38 (100\%) of flower poses were estimated correctly, despite the presence of occlusions (leaves partially covering some petals).
We attribute the high success rate to our filtering process using YOLO to mask the flower point cloud in preparation for point cloud registration.

\subsection{Microscope Subsystem Validations}
\subsubsection{Contact Planning and Detection}
Given a 6-D flower pose estimation, we hypothesize that the robot can 1) engulf the flower between the pollination tool and the microscope cup with the stem inside the fork opening, and 2) capture the flower on the concave cup support of the pollination tool.

\begin{table}[t]
\caption{Results from full pipeline deployment on 50 artificial flowers.}
\centering
\renewcommand{\arraystretch}{1.5}
\label{table: autonomous pipeline full results}
\begin{tabular}{l|ll|ll}
\multicolumn{1}{c|}{\textbf{Global}} & \multicolumn{2}{c|}{\textbf{Local}}                           & \multicolumn{2}{c}{\textbf{Micro}}                        \\ \hline
Detection                            & \multicolumn{1}{l|}{Alignment} & Pose Est.                    & \multicolumn{1}{l|}{Contact} & Inspection                 \\ \hline
\multicolumn{1}{r|}{80.0\%}          & \multicolumn{1}{r|}{95.0\%}    & \multicolumn{1}{r|}{100.0\%} & \multicolumn{1}{r|}{71.1\%}  & \multicolumn{1}{r}{98.2\%}
\end{tabular}
\end{table}

\textit{Experiment Setup:}
The robot starts from the aligned pose in front of the flower and executes the procedure in Sec.~\ref{subsubsec: contact planning and detection}.
A human observer visually inspects/confirms the success of engulfing the flower and capturing the flower on the cup support of the pollination tool.

\textit{Results and Discussion:}
We found that 38/38 (100\%) of the flowers were successfully engulfed, but only 27/38 (71.1\%) were captured on the tool cup.
The most common failure mode we noticed was the microscope failing to detect the flower center with HSV filtering when moving the tool upwards along the stem direction.
The second most common failure mode was due to obstacle (leaves and other stems).
To mitigate these issues, a key future direction is to explore motion planning around obstacles while closing the perception loop with both endoscope and microscope. 
We also note this experiment required a human observer to confirm success; one mitigation strategy could be incorporating tactile sensing on the pollination tool.

\subsubsection{Pollination Inspection}\label{subsubsec: pollination inspection experiment}
We hypothesize that HSV filtering as in Sec.~\ref{subsubsec: pollen deposition inspection} produces a significant increase in pixels classified as pollen from a microscope image when a flower is indeed pollinated.
We assume constant lighting, similar color from flower to flower, and flowers being positioned such that our microscope can autofocus.

\textit{Experiment Setup:}
Note, this experiment is \textit{not} run in the full system loop.
Instead, to ensure a controlled experiment, we use artificial strawberry flowers and commercially available talc powder as artificial strawberry pollen, because it has a similar average particle diameter of 20-30~$\mu$m~\cite{maas1977pollen,gilbert2018description}; see Fig.~\ref{fig: artificial plant and flower pollination detail}.
The microscope subassembly is separated from the robot upon a stationary platform.
Four artificial flowers are prepared with talc brushed across their fake stigma area, and four with no powder.
Each is placed under the microscope 7 times at different orientations (56 samples total), and the autofocus (Sec.~\ref{subsubsec: microscope focus score and autofocus}) and inspection procedures (Sec.~\ref{subsubsec: pollen deposition inspection}).
The system autonomous assesses the presence of pollen via the filtered pixel count being above a threshold.

\textit{Results and Discussion:}
For this test, 55/56 (98.2\%) of the images were correctly distinguished as either pollinated or not.
The one failure was a flower with talc powder applied more lightly than the others.
Excitingly, we also find that analyzing images in this way only takes seconds to execute compared to the minutes it takes for a human operator \cite{macinnis2017quantifying}, though our automation is not necessarily as accurate as an experienced human. 
We also note that for real flowers, distribution as well as quantity are important factors in successful pollination of strawberry plants, so developing a more sophisticated classifier may be needed.

\begin{table}[t]
\centering
\renewcommand{\arraystretch}{1.5}
\caption{Fruit yield from hand usage experiment with our pollination tool.}\label{table:pollination_result}
\begin{tabular}{l|r|r|r}
                         & \multicolumn{1}{l|}{\textbf{Rep 1}} & \multicolumn{1}{l|}{\textbf{Rep 2}} & \multicolumn{1}{l}{\textbf{Total}} \\ \hline
Fruit yield w/ our tool (human user)      & 582 g                               & 808 g                               & 1390 g                             \\
Fruit yield w/ artificial wind (control) & 451 g                               & 618 g                               & 1069 g                             \\
Increase in fruit mass with our tool   & 29.1\%                              & 30.7\%                              & 30.0\%                            
\end{tabular}
\end{table}

\begin{figure}[ht]
\centering
\includegraphics[width=1\columnwidth]{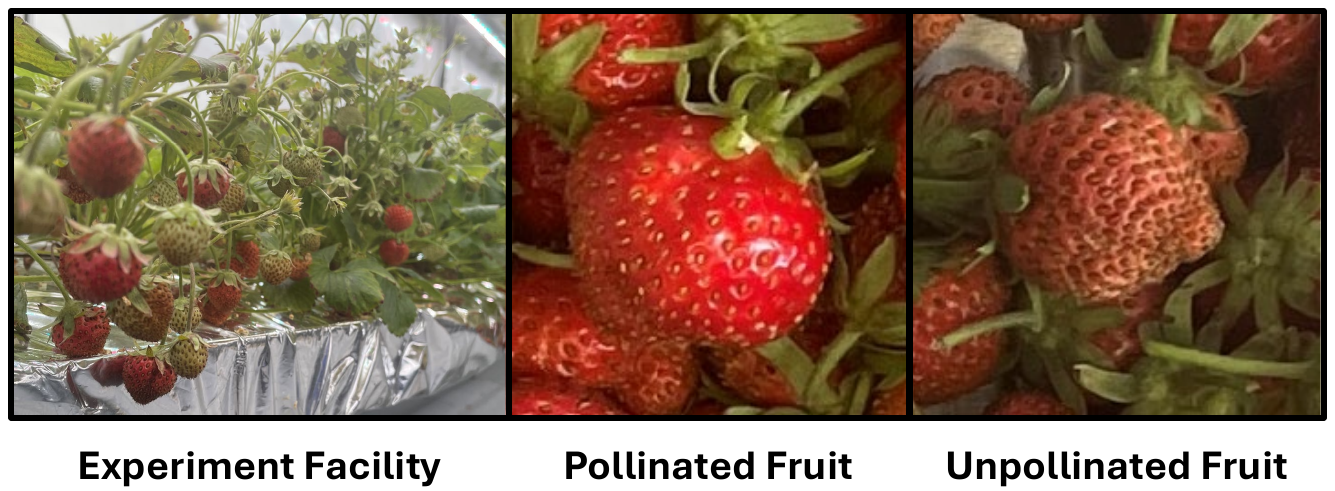}
\vspace{-1.5em}
\caption{
Buzz pollination tool validation results.
From left to right, the figures above shows the experiment facility (left), an example sufficiently pollinated fruit (middle), and an example insufficiently pollinated fruit (right).
The table below shows the fruit production results of the 64 strawberry plants. 
There was a $30\%$ increase in fruit mass for the plants pollinated with our tool.
}
\label{fig:pollination_exp}
\vspace*{-0.3cm}
\end{figure}

\subsubsection{Buzz Pollination}
We hypothesize that, by using our pollination tool, strawberry flowers will be sufficiently pollinated and yield better quality fruits than a control group of unpollinated strawberry flowers.

\textit{Experiment Setup:}
As an independent validation on the pollination tool, the experiment was conducted by hand in the USDA ARS facility in Charleston, NC.
We initialize the experiment on strawberry cultivar Mara Des Bois with all flowers and fruits removed on 08/13/2024. 
We perform two replications for both pollination and non-pollination group with a total of $2\regtext{ group}\times(2\regtext{ rep.}\times16\regtext{ plants})=64\regtext{ plants}$.
Note, the non-pollination group still passively receives some pollination due to a constant breeze in the facility from fans, which are needed to keep the plants moving so their leaves do not become damaged under artificial lighting.

The first pollination was conducted on 08/19/2024
The pollination tool was used for each newly developed flower with a vibration duration of 10 seconds. 
Pollination was applied every day on other newly-developed flowers. 
Mature red fruits were harvested for the first time on 09/13/2024.

\textit{Results and Discussion:}
As shown from Table \ref{table:pollination_result}, in both replications, the strawberry plants with pollination tool usage produced $30\%$ more fruit mass than those plants in the unpollinated group. 
Qualitatively, Fig.~\ref{fig:pollination_exp} shows representative examples of the fruit from both groups, indicating a significant increase in fruit quality as well.
The results indicate that our designed pollination tool is capable of sufficiently pollinating strawberry flowers.
Our next step is to run a fully automated experiment with the robotic system described in this paper, plus a comparison to hand brush pollination similar to \cite{nishimoto_evaluation_2023}.

\section{Conclusion}\label{sec:conclusion}

This paper proposed a robotic manipulator system that can autonomously perform \textit{and directly inspect} pollination of self-pollinating crops, specifically strawberries.
The system operates at three scales, from global (multi-flower) to local (single flower) to micro (pollen scale).
It uses a novel end effector design combining a vibrating pollination tool with an autofocusing mobile microscope.
The subsystems and overall pipeline were validated through extensive experiments on artificial and real flowers.
\textit{To the best of our knowledge, this is the first real-world autonomous robotic deployment of a handheld microscope in the wild} -- we are excited about the future possibilities of this technology.

However, the system is still limited in many ways; its flower detection would benefit from fine-tuning, the motion planning and perception rely on hand-tuned heuristics, the plants are not explicitly modeled as contact bodies, and the robot is not mounted on a mobile platform.
Future work will focus on extending to mobile manipulation and modeling robot-plant interactions.


    

\renewcommand{\bibfont}{\normalfont\footnotesize}
{\renewcommand{\markboth}[2]{}
\printbibliography}

\end{document}